\documentclass{article}

 \usepackage[preprint]{nips_2018}

\usepackage[utf8]{inputenc} 
\usepackage[T1]{fontenc}    
\usepackage{hyperref}       
\usepackage{url}            
\usepackage{booktabs}       
\usepackage{amsfonts}       
\usepackage{nicefrac}       
\usepackage{microtype}      

\usepackage{amsfonts,amssymb,amsmath,amsthm} 
\usepackage{natbib}
\usepackage{mathrsfs}
\newcommand{\R}{\mathbb{R}}
\newcommand{\SigS}{\mathscr{S}}
\newcommand{\LipSet}{$\SigS_{(A^{-1},\omega)}$}
\newcommand{\Q}{\textit{Q.E.D.}}

\newcommand{\Ge}{G_\epsilon}
\newcommand{\Ac}{A\circ}

\title{Lipschitz Learning for Signal Recovery}

\author{
  Hong Jiang, Jong-Hoon Ahn and Xiaoyang Wang
  \\
  Nokia Bell Labs\\
  Murray Hill, NJ 07974\\
  \texttt{\{hong.jiang,jong\_hoon.ahn,xiaoyang.wang\}@nokia-bell-labs.com} \\
}

\begin{document}

\maketitle

\begin{abstract}
  
  We consider the recovery of signals from their observations, which are samples of a transform of the signals rather than the signals themselves, by using machine learning (ML).
  We will develop a theoretical framework to characterize the signals that can be robustly recovered from their observations by an ML algorithm, and establish a Lipschitz condition on signals and observations that is both necessary and sufficient for the existence of a robust recovery.
  We will compare the Lipschitz condition with the well-known restricted isometry property of the sparse recovery of compressive sensing, and show the former is more general and less restrictive.
  For linear observations, our work also suggests an ML method in which the output space is reduced to the lowest possible dimension.
  
\end{abstract}

\section{Introduction}\label{intro}
 In many applications, a signal is only available as observations which are samples of a transform of the signal, rather than the samples of the signal itself. Examples are compressive sensing 
 \cite{candes2006,candes2005}, 
 in which a signal is under a dimension-reducing linear transform, wireless communications in which a signal undergoes a linear or nonlinear channel transform \cite{kim,jiang}, and  computational imaging in which the acquired data is a result of light field going through a transform due to optical devices \cite{duarte,sun20133d}.
 Recently, machine learning (ML) algorithms have demonstrated superior performance in recovering signals from observations  \cite{goodfellow,mousavi,kulkarni}. To recover a signal from its observation, an ML model, such as a convolutional neural network, is trained so that the recovered signal is the output of the model when its observation is used as the input. 
 Despite the great success of ML in recovering signals from observations and the development in ML theory in general \cite{valiant, vapnik, blumer, kearns, AbuMostafa2012}, there is a lack of theoretical understanding in many aspects of ML recovery. 
 
 This work is to address the question of under what condition a signal can be recovered from its observation by an ML algorithm. We develop a theoretical framework to characterize the signals that can be robustly recovered from observations.
 We will establish a Lipschitz condition on signals and observations and show that it is both necessary and sufficient for the existence of a robust ML algorithm to recover the signals. We will compare the Lipschitz condition with the restricted isometry property (RIP)  \cite{candes2006,candes2005} in the sparse signal recovery of compressive sensing, and show that the former is more general and less restrictive.
 
 The set of signals satisfying the Lipschitz condition is not unique. Since there is no restriction on the transform of the observations, there is no expectation that a given set of signals can be recovered from their observations. Instead, what is expected is that all signals with certain structure should be recovered. In our framework, the structure of the recoverable signals is precisely defined by the Lipschitz condition: each Lipschitz set is a set of structured signals that are robustly recoverable and a different set defines a different structure. A finite number of training signals can always be used to define a Lipschitz set of signals which can be robustly recovered by a trained, robust ML model.
 
The significance of this work is that it not only answers the theoretical question of what signals can be robustly recovered, but also suggests a practical recovery method by using singular value decomposition (SVD) for linear observations (see Theorem~3), in which the dimension of the output space of the target function is reduced to the minimum possible.

All proofs of the paper will be given in Appendix.


 
\section{Related Work}\label{related}
 
 The term ``Lipshitz learning’’ was previously used for classification on graphs \cite{kyng}, in which the target functions in graph-based semi-supervised learning are Lipschitz. In this paper, the same term is used in a different context and a broader sense. In this paper, Lipschitz learning refers to the framework of recovering signals satisfying the Lipschitz condition. Since in this work, the recovery can be achieved by Lipschitz hypothesis, the use of the term here is consistent with its previous use.
 
 In addition to \cite{kyng}, existing work in \cite{Luxburg,koltchinskii,lopez-paz} also studies to utilize Lipschitz functions as decision functions or target functions for the classification problem. Specifically, \cite{Luxburg} finds that the Lipschitz function is a generalization of decision functions for metric spaces, and shows that several well-known algorithms are special cases of the Lipschitz classifier. \cite{koltchinskii} poses the cause-effect inference problem as a classification problem, and uses the property of Lipschitz function to derive the bound on excess risk. In addition, Lipschitz function is used in \cite{lopez-paz} for theoretic analysis of empiric risk minimization. Our work differs from the existing work \cite{Luxburg,koltchinskii,kyng,lopez-paz} in the following two aspects: 1) We utilize the Lipschitz condition for the problem of general signal recovery, whereas \cite{Luxburg,koltchinskii,kyng,lopez-paz} utilizes Lipschitz functions for the problem of classification. 2) To the best of our knowledge, no existing work utilizes the property of Lipschitz set, which is essential in our theory of signal recovery with Lipschitz learning.
 
 Our framework shows that the Lipschitz condition on a set of signals is equivalent to the existence of a hypothesis for the recovery of these signals. It differs from the probably approximate correct (PAC) learning \cite{valiant}, and the statistical learning theory \cite{vapnik} that analyzes the probability in successfully finding a hypothesis with low generalization error. 
Our work is currently concerned with the existence of Lipschitz hypothesis, but in the future, will address the complexity of Lipschitz learning such as reducing the bound on number of total training samples required, which includes, for example, using a probabilistic model in Lipschitz learning.

\section{Lipschitz Learning}\label{lipschitz}
\textbf{Problem Definition.}
Let $x\in\mathbb{R}^N$ be a signal, $A:\mathbb{R}^N\to\mathbb{R}^M$ be an operator with $M\leq N$. The observation of signal $x$ under transform $A$ is $y=A\circ x \in \mathbb{R}^M$, where the symbol "$\circ$" means "operates on". The operator $A$ may be linear or nonlinear, and it may not be an injection even when $M=N$. The objective here is, for a given $A$, to recover the signal $x$ from its observation $y$ by a machine learning algorithm. In an ML algorithm, a hypothesis is a computable function $G:\mathbb{R}^M\to\mathbb{R}^N$. A recovered signal $\tilde{x}$ from the observation $y$ by the hypothesis $G$ is $\tilde{x}= G(y)=G(A\circ x)$, with $\tilde{x}\approx x$. 

Since $A$ may not be injective, there is no expectation that a signal can be uniquely recovered from a given observation $y\in A\circ \R^M$. Instead, we attempt to characterize a set of signals that can be robustly recovered from their observations by an ML algorithm. Such a characterization is tantamount to imposing a structure on signals to ensure the success of  recovery. For example, in compressive sensing \cite{candes2006,candes2005}, observations are the results of a singular linear transform but it is possible to uniquely recover a set of sparse signals under certain conditions.

Let $\SigS\subseteq\R^N$ be a set of signals. For all signals in $\SigS$ to be recovered from their observations, a necessary condition is 
\begin{equation}
\textrm{for all }x^1,x^2\in\SigS, \|x^1 - x^2\| = 0, \textrm{ if } \| A\circ x^1 - A\circ x^2\| = 0. 
\label{recoverable}
\end{equation}
Furthermore, for a recovery to be robust and resilient to noise, it is required that
\begin{equation}
\textrm{for all }x^1,x^2\in\SigS, \|x^1 - x^2\| \sim \textrm{small}, \textrm{ if } \| A\circ x^1 - A\circ x^2\| \sim \textrm{small}.
\label{robust}
\end{equation}

Motivated by Eqs.~\eqref{recoverable} and \eqref{robust}, we make the following definition.

\medskip
\textbf{Definition 1.}
Given $\omega>0$, a set $\SigS\subseteq\R^N$ is said to be $(A^{-1},\omega)\textrm{-Lipschitz}$ if  
\begin{equation}
\|x^1 - x^2\| \leq \omega \| A\circ x^1 - A\circ x^2\|, \textrm{for all }x^1,x^2\in\SigS.
\label{LipSetEq}
\end{equation}
A set is said to be a Lipschitz set if there is an $\omega>0$ such that it is 
$(A^{-1},\omega)$-Lipschitz. We denote an $(A^{-1},\omega)\textrm{-Lipschitz}$ set by $\SigS_{(A^{-1},\omega)}$, and call the signals in a Lipschitz set the Lipschitz signals.

\medskip
Note in Definition~1, $A^{-1}$ is simply a notation; it doesn't mean $A^{-1}$ exists. However, when restricted on \LipSet, $A|$\LipSet~ does have an inverse, and its inverse is Lipschitz. An ML algorithm is to find a hypothesis to approximate it.

The Lipschitz condition in Eq.~\eqref{LipSetEq} is a joint condition on signals and their observations (or the operator $A$). It may be framed in the following two ways.

\hspace{10pt} 1) For a given set of signals $\SigS$,  Eq.~\eqref{LipSetEq} is a condition on the operator $A$. It is equivalent to saying that the inverse $A^{-1}$ must exist on $\SigS$, and the inverse is $\omega$-Lipschitz. Traditional signal recovery algorithms, such as $\ell_1$ minimization in compressive sensing, is within this framework, i.e., they attempt to recover all signals in a given structure under the assumption that the operator $A$ meets certain conditions.

\hspace{10pt} 2) For a given operator $A$, Eq.~\eqref{LipSetEq} is a condition on a set of signals $\SigS$ to be recovered. For any operator $A$, there is always a set satisfying Eq.~\eqref{LipSetEq}: any singleton set. An ML algorithm may be designed to recover those signals of interest that are recoverable for the given $A$, by properly selecting training signals to define a set of signals of interest to satisfy Eq.~\eqref{LipSetEq}. In this context, the Lipschitz signals are the structured signals. 

\textbf{Example.} Let operator $A$ be the continuous function $[0,~3]\to\R$ defined by
\begin{equation}
A(x)\triangleq
\begin{cases}
x, &\quad\text{if }x\in[0,~1)\\
1, &\quad\text{if }x\in[1,~2]\\
x-1,  &\quad\text{if }x\in(2,~3].\\
\end{cases}
\end{equation}
The set $[0,3]$ is not Lipschitz, but $[0,~1]$, or $[2,~3]$, is an $(A^{-1},1)$-Lipschitz set.
For any $\alpha\in[1,~2]$, the set $[0,~1)\cup\{\alpha\}\cup(2,~3]$ is not Lipschitz although $A$ is injective on it; clearly, signals in $[0,~1)\cup\{\alpha\}\cup(2,~3]$ cannot be recovered reliably under noise because a small noise in the observation $y=A(x)=1$ may cause the recovered signal to be $\tilde{x}<1$ or $\tilde{x}=\alpha$ or $\tilde{x}>2$. On the other hand, $[0,~1-\epsilon_1]\cup\{\alpha\}\cup[2+\epsilon_2,~3]$ is a Lipschitz set for any $\epsilon_1, \epsilon_2>0$. For example, for any $\omega>1$,
$[0,~1-\frac1{2(\omega-1)}]\cup\{\frac32\}\cup[2+\frac1{2(\omega-1)},~3]$
 is $(A^{-1},\omega)$-Lipschitz, and so is $[0,~1]\cup[2+\frac1{\omega-1},~3]$ .

\medskip

\textbf{Property 1.} If $\Omega\subseteq\R^N$ is a finite set, i.e., $|\Omega|<+\infty$, and $A$ is an injection on $\Omega$, then $\Omega$ is $(A^{-1}, \omega)$-Lipschitz where 
\begin{equation}
\omega \triangleq \underset{x^1,x^2\in\Omega, x^1\neq x^2}{\textrm{max}} \frac{\|x^1-x^2\|}{\|A\circ x^1 - A\circ x^2\|}.
\label{finiteomega}
\end{equation}

\textbf{Property 2.} Let $A$ be a linear operator,  and \LipSet~ be an $(A^{-1},\omega)$-Lipschitz set. Then any scaled and shifted set from \LipSet~ is also an $(A^{-1},\omega)$-Lipschitz set. More precisely, for any $\alpha\in\R$, and $s\in\R^N$, 
\begin{equation}
\alpha \SigS_{(A^{-1},\omega)}+s \triangleq \{\alpha x+s ~|~x\in \SigS_{(A^{-1},\omega)}\}
\end{equation}
is  an $(A^{-1},\omega)$-Lipschitz set.

\medskip
Property~1 shows that any finite set on which $A$ is injective is Lipschitz, and therefore, it can be used as a starting point to build a Lipschitz set of signals of interest. For example, a finite set of training signals may be used to define a maximal set of Lipschitz signals that includes the training signals.

  \medskip
\textbf{Definition 2.}
A machine learning hypothesis $G: \R^M\to\R^N$ is said to be $\omega$-Lipschitz, for $\omega>0$, if 
\begin{equation}
\|G(y^1)-G(y^2)\|\leq\omega\|y^1-y^2\|, \textrm{ for all } 
y^1,y^2\in\R^M.
\end{equation}
A hypothesis $G$ is said to be Lipschitz, or robust, if there is an $\omega>0$ such that $G$ is $\omega$-Lipschitz.

\medskip
\textbf{Definition 3.}
A set $\Omega$ is said to be labeled if every $x\in\Omega$ and its observation $y=A\circ x$ are known.

\section{Characterization of Signal Recovery}\label{characterization}

In this section, we will show the Lipschitz condition on a set of signals is equivalent to the existence robust ML hypothesis for recovery of the signals. More precisely, we will show that the Lipschitz condition Eq.~\eqref{LipSetEq} is both necessary and sufficient for the existence of Lipschitz hypotheses in the ML signal recovery.

In the rest of this paper, we assume the observations are bounded, which is generally the case in practice. Without loss of generality, we may assume they are bounded by the unit hypercube, i.e., 
\begin{equation}
\{ A\circ x ~|~x\in\R^{N}\}\subseteq [0,1]^M \subset \R^M .
\label{bounded}
\end{equation}

\medskip
\textbf{Lemma 1.}
\textit{ 
Let $\Omega\subseteq\R^N$ be a finite set and labeled, and $A:\Omega\to\R^M$ be an injection. Then there exists an $\omega$-Lipschitz hypothesis $G:\R^M\to\R^N$ such that }
\begin{equation}
G(A\circ x)= x, \textit{ for all } x\in\Omega.
\label{training}
\end{equation}

Lemma~1 is an application of the McShane-Whitney extension theorem \cite{mcshane,whitney}. It provides an explicit and constructive Lipschitz hypothesis on a finite labeled set (see proof in Appendix). Furthermore, Eq.~\eqref{training} shows that the finite set is a training set for the Lipschitz hypothesis. The training set can then be expanded to a Lipschitz set in which all signals can be recovered robustly, as to be seen in the next Theorem which shows that the Lipschitz set is sufficient for the existence of a Lipschitz hypothesis.

\medskip
\textbf{Theorem 1.}
\textit{ 
	Let \LipSet $\subseteq\R^N$ be an $(A^{-1},\omega)$-Lipschitz set. Then for any $\epsilon>0$, there exists a finite set $\Omega_{A,\omega,\epsilon}\subseteq$\LipSet. If $\Omega_{A,\omega,\epsilon}$ is labeled, then there exists an $\omega \sqrt{N}$-Lipschitz hypothesis $G_\epsilon:\R^M\to\R^N$, such that} 
	
	\hspace{10pt} \textit { (i) ~$G_\epsilon(A\circ x)=x$, for all $x\in\Omega_{A,\omega,\epsilon}$; (Training)}
	
	\hspace{10pt} \textit { (ii)  $\|G_\epsilon(A\circ x)-x\|\leq \epsilon$, for all $x\in$\LipSet. (Recovery of all $(A^{-1},\omega)$-Lipschitz signals)
}

\medskip
The factor $\sqrt{N}$ in $\omega\sqrt{N}$-Lipschitz is not necessary and can be removed; it is there only to simplify the proof, which is given in Appendix.

Theorem 1 means that if a set of signals is Lipschitz, then for any given precision, there exists a finite set of training signals so that a Lipschitz hypothesis can be trained on the finite set to recover all signals within the given precision. It guarantees that a set of Lipschitz signals can be recovered by a robust ML algorithm, to an arbitrary precision. We note that although the training set $\Omega_{A,\omega,\epsilon}$ in Theorem~1 is finite in theory, it may be too large for practical purposes.

A Lipschitz hypothesis is stronger than a continuous target function. It could be argued that a continuous target function is sufficient to provide robustness of recovery, so a question would arise as to if the Lipschitz condition \eqref{LipSetEq} is too strong. However, since a set of signals $\SigS$ may be discrete, not continuous or connected, it is not possible to define a "continuity" on $\SigS$ in the classic sense to guarantee robust recovery, as Lipschitz condition \eqref{LipSetEq} does. The Lipschitz set in Definition~1 is a sensible condition on a (possibly discrete) set of signals for robust recovery.
More discussions regarding the Lipschitz condition will be given in Section~\ref{comparision}.

Next, we show that the Lipschitz set is necessary for the existence of Lipschitz hypothesis.

\medskip
\textbf{Theorem 2.}
\textit{ 
	Let $\SigS\subseteq\R^N$ be a set. If there exists $\omega > 0$ such that for any $\epsilon>0$ there is an $\omega$-Lipschitz hypothesis $G_\epsilon:\R^M\to\R^N$, such that }
\begin{equation}
\|G_\epsilon(x)-x\|\leq\epsilon \textrm{ for all }x\in\SigS,
\label{recovered}
\end{equation}
\textit{then $\SigS$ is an $(A^{-1},\omega)$-Lipschitz set.
}

\medskip
Theorem 2 says that if there are $\omega$-Lipschitz hypotheses to recover a set of signals to an arbitrary precision, then the set itself must be $(A^{-1},\omega)$-Lipschitz. A weaker version is given below.

\medskip
\textbf{Corollary 1.}
\textit{ 
	Let $\SigS\subseteq\R^N$ be a set. If there exist an $\epsilon>0$ and an $\omega$-Lipschitz hypothesis $G_\epsilon:\R^M\to\R^N$  such that $\|G_\epsilon(x)-x\|\leq\epsilon$ for all $x\in\SigS$, then $\SigS$ satisfies
	 }
\begin{equation}
\|x^1-x^2\|\leq 2\epsilon + \omega \|A\circ x^1-A\circ x^2\|,\textrm{ for all } x^1,x^2\in\SigS.
\label{weakerLip}
\end{equation}

\medskip
This result says that if a set of signals can be recovered to a certain precision by a Lipschitz hypothesis, then the set of signals is approximately Lipschitz, up to the precision of the recovery.

 Theorems 1 and 2 completely characterize robust ML signal recovery: a set of signals can be robustly recovered by  ML algorithms if and only if the set satisfies the Lipschitz condition \eqref{LipSetEq}.

For linear operators, we have a stronger version of Theorem~1 as follows.

\medskip
\textbf{Theorem 3.}
\textit{ 
	Let $A$ be linear, and \LipSet $\subseteq\R^N$ be an $(A^{-1},\omega)$-Lipschitz set. Then there exist matrices $\Psi\in\R^{M\times M}$ and $V\in\R^{N\times N}$.
	Furthermore, for any $\epsilon>0$, there exists a finite set $\Omega_{A,\omega,\epsilon}\subseteq$\LipSet. If $\Omega_{A,\omega,\epsilon}$ is labeled, then there exists an $\omega \sqrt{N-M}$-Lipschitz hypothesis $G_\epsilon:\R^M\to\R^{N-M}$, such that
	the mapping $R:\R^M\to\R^N$ defined by $ R(y) \triangleq V 
	\left [ \begin{array}{c}
	\Psi y \\
	G_\epsilon(y)\\
	\end{array}
	\right ]
	$ satisfies
	} 

\hspace{10pt} \textit { (i) ~~$R(A x)=x$, for all $x\in\Omega_{A,\omega,\epsilon}$; (Training)}

\hspace{10pt} \textit { (ii) ~$\|R(A x)-x\|\leq \epsilon$, for all $x\in$\LipSet;  (Recovery of all $(A^{-1},\omega)$-Lipschitz signals)
}

\hspace{10pt} \textit { (iii) $A(R(A x))=Ax$, for all $x\in\R^N$. (Recovered signals match the observations)
}

\medskip

The significance of Theorem 3 as compared to Theorem 1 is twofold. First, the output space of the hypothesis $G_\epsilon$ in Theorem~3 has lower dimension than that of Theorem~1: $\R^{N-M}$ in Theorem~3 vs $\R^N$ in Theorem~1. Consequently, the bound on the total number of required training signals is lower in Theorem~3 than in Theorem~1. Secondly, the recovered signals have the same observations as the original signals, as in (iii).
In other words, even if the recovered signal $\tilde{x}=R(Ax)$ may not equal the original signal $x$, the observations are the same: $A\tilde{x}=AR(Ax)=Ax$, i.e., the recovered signal $\tilde{x}=R(Ax)$ is indistinguishable from the original signal $x$ in the observation space.

\section{Comparison with Sparse Recovery in Compressive Sensing}\label{comparision}

In this section, we assume the operator $A$ is linear, i.e., $A\in\R^{M\times N}$.

\medskip
\textbf{Sparse recovery \cite{candes2006}}

Let $S\leq N$, $T\subset\{1,...,N\}$, and $A_T$ be the $M\times |T |$ submatrix obtained by extracting the columns of $A$ corresponding to the indices in $T$. $A$ is said to satisfy $S$-restricted isometry property (RIP) if there exists $\delta_S\in(0,1)$ such that
\begin{equation}
(1-\delta_S)\|c\|^2\leq\|A_T\cdot c\|^2 \leq (1+\delta_S)\|c\|^2,
\label{RIP}
\end{equation}
for all subsets $T$ with $|T| \leq S$ and coefficient sequences $(c_j )_{j\in T} $. $\delta_S$ is said to be $S$-restricted isometry constant. 
It is shown in \cite{candes2005} that if $A$ satisfies the RIP with 
\begin{equation}
\delta_{2S}+\delta_{3S}<1,
\label{RIPconst}
\end{equation}
then an $S$-sparse signal $x$ can be recovered from its observation $y=Ax$ by $\ell_1$-minimization \cite{candes2006}.

\medskip
\textbf{Sparse signals with RIP conditions \eqref{RIP} and \eqref{RIPconst} are Lipschitz}

Let the set of $S$-sparse signals be
$\SigS_S \triangleq\{x\in\R^N~|~x \textrm{ is $S$-sparse}\}$. 
Conditions \eqref{RIP} and \eqref{RIPconst} in fact imply that $\SigS_S$ is $(A^{-1},\frac1{\sqrt{1-\delta_{2S}}})$-Lipschitz. 
Indeed, let $x^1,x^2\in\SigS_S$. Then $x^1-x^2$ is $2S$-sparse. According to Eq.~\eqref{RIP}, 
\begin{equation}
(1-\delta_{2S})\|x^1-x^2\|^2\leq\|A\cdot (x^1-x^2)\|^2 ,
\label{RIP2S}
\end{equation}
where $\delta_{2S}<1$ according to \eqref{RIPconst}. Condition \eqref{RIP2S} leads to
\begin{equation}
\|x^1-x^2\|\leq\frac1{\sqrt{1-\delta_{2S}}}\|A\cdot (x^1-x^2)\|= \frac1{\sqrt{1-\delta_{2S}}}\|A x^1-A x^2\|.
\end{equation}
Therefore, the set of $S$-sparse signals for which RIP  with \eqref{RIPconst} is satisfied is an $(A^{-1},\frac1{\sqrt{1-\delta_{2S}}})$-Lipschitz set, and consequently, according to Theorem~1, there exists a robust ML recovery algorithm for the $S$-sparse signals if RIP is satisfied with condition \eqref{RIPconst}.

This shows that the Lipschitz condition \eqref{LipSetEq} is more general and less restrictive than the RIP conditions \eqref{RIP} and \eqref{RIPconst}. Of course, it must also be pointed out that the stronger RIP conditions \eqref{RIP} and \eqref{RIPconst} lead to a strong and constructive result that $S$-sparse signals can be recovered by $\ell_1$-minimization.

\section{Conclusion}
We have developed a framework to characterize the robust ML signal recovery. The theory  in the framework makes the terminology "structured signals" in traditional signal recovery algorithms more precise. Here, the structured signals are the Lipschitz signals. For any given transformation $A$, it is always possible to define a set of Lipschitz signals, i.e., structured signals, so that they can be robustly recovered by a trained ML model.

Although we have provided a complete characterization of ML signal recovery in theory, more work is needed to render this theoretical framework for practical use in general. For example, the bound on the total number of training signals required to guarantee robust recovery in this framework is too high to be used in practice. However, this theoretical work does provide insights that can guide the design of practical ML signal recovery algorithms. For linear systems, Theorem~3  suggests a practical method of using SVD to reduce the dimension of the output space of an ML model from $\R^N$ to $\R^{N-M}$, which is the minimum possible dimension on which a recovery algorithm must learn.

\section*{Appendix}

\textbf{Proof of Property 1.}
If $\Omega\subseteq\R^N$ is a finite set and $A$ is injective on $\Omega$, then $\omega$ is well-defined in \eqref{finiteomega}, and furthermore,
\begin{equation}
\frac{\|x^1-x^2\|}{\|A\circ x^1-A \circ x^2\|}\leq\omega, \textrm{ for all } x^1,x^2\in\Omega \textrm{ with } x^1\neq x^2,
\end{equation}
which shows $\|x^1-x^2\|\leq \omega\|A\circ x^1-A \circ x^2\|$, i.e., $\Omega$ is an $(A^{-1},\omega)$-Lipschitz set. \Q

\textbf{Proof of Property 2.} Let $u^1,u^2\in \alpha$\LipSet$+s$. There exist $x^1,x^2\in$\LipSet,
 with $u^i=\alpha x^i+s,i=1,2$, so
\begin{align*}
\|u^1-u^2\|=\|\alpha x^1 +s - (\alpha x^2+s)\| = |\alpha|\|x^1-x^2\|\overset{x^1,x^2\in\textrm{\LipSet}}{\leq} &
|\alpha|\omega \|A\circ x^1-A \circ x^2\| \\
= ~~~~~~~~~&\omega \|\alpha( A\circ x^1-A \circ x^2)\| \\
\overset{A\textrm{ is linear }}= ~~~~&\omega \|A\circ u^1-A \circ u^2\|,
\end{align*}
which shows that $\alpha$\LipSet$+s$ is an $(A^{-1},\omega)$-Lipschitz set. \Q

\textbf{Proof of Lemma 1.}
Since $\Omega$ is finite and $A$ is injective on it, it follows from Property~1 that $\Omega$ is $(A^{-1},\omega_1)$-Lipschitz for some $\omega_1>0$. Following McShane-Whitney extension theorem \cite{mcshane,whitney}, we define $g_i: \R^M\to\R$ by
\begin{equation}
g_i(y)\triangleq\underset{x\in\Omega}{\min}\{x_i+\omega_1\|y-A\circ x\|\}, y\in\R^M, i=1,...,N.
\label{MWET}
\end{equation}
We show that $g_i$ is $\omega_1$-Lipschitz. Indeed, since $\Omega$ is finite, for any $y^1,y^2\in\R^M$, there exists $x^2\in\Omega$ such that $g_i(y^2)=x^2_i+\omega_1\|y^2-A\circ x^2\|$. Furthermore, from definition \eqref{MWET}, $g_i(y^1)\leq x^2_i+\omega_1\|y^1-A\circ x^2\|$, and therefore,
\begin{equation}
\begin{split}
g_i(y^1)-g_i(y^2)&=g_i(y^1)-\left ( x^2_i+\omega_1\|y^2-A\circ x^2\| \right ) \\
&\leq  x^2_i+\omega_1\|y^1-A\circ x^2\|  -\left ( x^2_i+\omega_1\|y^2-A\circ x^2\| \right ) \\
&= \omega_1 \left (\|y^1-A\circ x^2\| -\|y^2-A\circ x^2\| \right )\\
&\leq \omega_1 \left \| (y^1-A\circ x^2) - (y^2-A\circ x^2) \right \|
=\omega_1\|y^1- y^2\|.
\end{split}
\label{LipOmegagi1}
\end{equation}
Reversing the roles of $y^1$ and $y^2$ in \eqref{LipOmegagi1}, we also have $g_i(y^2)-g_i(y^1) \leq  \omega_1\|y^1- y^2\|$. This, together with \eqref{LipOmegagi1}, shows
\begin{equation}
|g_i(y^1)-g_i(y^2)| \leq  \omega_1\|y^1- y^2\|.
\label{LipOmegagi2}
\end{equation}
Define $G:\R^M\to\R^N$ by
\begin{equation}
G(y)\triangleq\left [
\begin{array}{c}
g_1(y)\\ \vdots \\ g_N(y)
\end{array}
\right ].
\end{equation}
Let $\omega\triangleq\omega_1\sqrt N$. From \eqref{LipOmegagi2}, we have
\begin{equation}
\|G(y^1)-G(y^2)\|
=\sqrt{\sum_{i=1}^{N} |g_i(y^1)-g_i(y^2)|^2}
\overset{\textrm{\eqref{LipOmegagi2}}}{\leq} 
\omega_1\sqrt{N}\|y^1-y^2\|=\omega\|y^1-y^2\|,
\label{LipOmegaN}
\end{equation}
which shows  $G$ is $\omega$-Lipschitz. From 
\eqref{MWET}, it's easy to verify that $G(A\circ x)=x$. \Q ~~It is important to point out that Eq.\eqref{MWET} provides an explicit and constructive Lipschitz hypothesis.

\textbf{Proof of Theorem 1.}
By assumption, $A\circ$\LipSet$\subseteq[0,1]^M$. Now define 
\begin{equation}
t\triangleq\left \lceil \frac{1+\sqrt{N}}{\epsilon}\omega\sqrt{M} \right \rceil;
\Delta_j\triangleq\left [\frac{j-1}{t},\frac{j}{t} \right ], j=1,\cdots,t;
[0,1]^M=\underset{j=1}{\overset{t^M}{\cup}}v_j, 
v_j\triangleq\overset{M}{\underset{k=1}{\times}} \Delta_{j_k}.
\label{hypercubes}
\end{equation}
Each $v_j$ in \eqref{hypercubes} is a hypercube in $\R^M$ of length $\frac1t$ in each dimension, and therefore,
we have
\begin{equation}
\|y^1-y^2\|\leq \frac{\sqrt{M}}{t}\leq \frac{\epsilon}{\omega(1+\sqrt N)},
\textrm{ for all } 
y^1,y^2\in v_j.
\label{incube}
\end{equation}
We now define
\begin{equation}
\Omega_{A,\omega,\epsilon}\triangleq\overset{t^M}{\underset{j=1}{\cup}}\Omega^j, \textrm{ where } \Omega^j\triangleq
\begin{cases}
	\{x^j\},&\textrm{if }\exists~x^j\in\textrm{\LipSet}\backepsilon A\circ x^j\in v_j \\
		\varnothing, & \textrm{otherwise }\\
\end{cases}.
\label{selection}
\end{equation}
It is clear that $\Omega_{A,\omega,\epsilon}\subseteq$\LipSet ~is $(A^{-1},\omega)$-Lipschitz and finite with 
\begin{equation}
|\Omega_{A,\omega,\epsilon}| \leq t^M = \left \lceil \frac{1+\sqrt{N}}{\epsilon}\omega\sqrt{M} \right \rceil^M.
\label{Obound}
\end{equation}
It follows from Lemma~1 that there exists an $\omega\sqrt N$-Lipschitz hypothesis $G_\epsilon:\R^M\to\R^N$ and $G_\epsilon(A\circ x)=x$ for all $x\in \Omega_{A,\omega,\epsilon}$, which proves (i).

We now show (ii). For any $x\in$\LipSet, because $A\circ x\in A\circ$\LipSet$\subseteq[0,1]^M=\underset{j=1}{\overset{t^M}{\cup}}v_j$, according to \eqref{bounded} and \eqref{hypercubes}, there exists a $j$ such that $A\circ x\in v_j$. From \eqref{selection}, $\Omega^j=\{x^j\}$ and $A\circ x^j\in v_j$.
\begin{equation}
\begin{aligned}
\|G_\epsilon(A\circ x)-x\|\hspace{50pt} \leq & ~~\|G_\epsilon(A\circ x)-x^j\|+\|x^j-x\|\\
\overset{x^j\in\Omega_{A,\omega,\epsilon}\textrm{ and (i)}}{=}&~~\|G_\epsilon(A\circ x)-G_\epsilon(A\circ x^j)\|+\|x-x^j\|\\
\overset{G_\epsilon\textrm{ is }\omega\sqrt{N}\textrm{-Lipschitz}}{\leq}&~~\omega\sqrt{N}\|A\circ x-A\circ x^j\|+\|x-x^j\|\\
\overset{x,x^j\in\textrm{\LipSet}}{\leq}&~~\omega\sqrt{N}\|A\circ x-A\circ x^j\|+
\omega\|A\circ x-A\circ x^j\|\\
\overset{A\circ x,A\circ x^j\in v_j\textrm{ and \eqref{incube}}}{\leq} &~~\omega(1+\sqrt N)\frac{\epsilon}{\omega(1+\sqrt N)}=\epsilon. ~~~~~~~~~~~~~~~~~~~~~ Q.E.D.
\end{aligned}
\end{equation}

\textbf{Proof of Theorem 2.}
To show $\SigS$ is $(A^{-1},\omega)$-Lipschitz, let $x^1,x^2\in\SigS$. For any $\epsilon>0$, we have
\begin{equation}
\begin{aligned}
\|x^1-x^2\|=&\|x^1-G_\epsilon(A\circ x^1)+\Ge(\Ac x^2)-x^2+\Ge(\Ac x^1)-\Ge(\Ac x^2)\| \\
\leq &\|x^1-G_\epsilon(A\circ x^1)\|+\|\Ge(\Ac x^2)-x^2\|+\|\Ge(\Ac x^1)-\Ge(\Ac x^2)\| \\
\overset{\textrm{\eqref{recovered}}}{\leq} & 2\epsilon +\|\Ge(\Ac x^1)-\Ge(\Ac x^2)\|
\overset{\Ge\textrm{ is }\omega\textrm{-Lipschitz}}{\leq} 
 2\epsilon +\omega\|\Ac x^1-\Ac x^2\|. \\
\end{aligned}
\label{Proof2}
\end{equation}
Since $\epsilon$ is arbitrary while other variables are fixed, \eqref{Proof2} implies 
\eqref{LipSetEq}, i.e., $\SigS$ is Lipschitz.
\Q

\textbf{Proof of Corollary 1.} 
Eq.~\eqref{weakerLip} of Corollary 1. follows immediately from \eqref{Proof2}. \Q

\textbf{Proof of Theorem 3.} We start by following the same process as in the proof of Theorem 1, but change the factor $\sqrt{N}$ in \eqref{hypercubes} to $\sqrt{N-M}$ to obtain hypercubes $v_j$ and a finite set $\Omega_{A,\omega,\epsilon}\subseteq $\LipSet. Instead of $t$ defined in \eqref{hypercubes} and the bounds  derived in \eqref{Obound} and \eqref{incube}, we now have
\begin{equation}
Q\triangleq 1+\sqrt{N-M};~
t\triangleq\left \lceil \frac{\omega Q}{\epsilon}\sqrt{M} \right \rceil;~
|\Omega_{A,\omega,\epsilon}| \leq 
t^M  
;~
\|y^1-y^2\|\leq \frac{\epsilon}{\omega Q},
\textrm{ for all } 
y^1,y^2\in v_j.
\label{thm3}
\end{equation}

Without loss of generality, we assume $A\in\R^{M\times N}$ is full rank (if not, $M$ can be reduced until it is). Performing the singular value decomposition (SVD) of $A$, we have 
\begin{equation}
A=U[\Sigma ~~~\textbf{0}]V^T; ~
 U,\Sigma\in \R^{M\times M}, 
  ~ \Sigma=\textrm{diag}\{\sigma_1,\cdots,\sigma_M\};
~ \textbf{0}\in\R^{M\times(N-M)},V\in\R^{N\times N}.
\end{equation}
$U,V$ are unitary matrices, and \textbf{0} is the matrix with all entries being 0. $\sigma_j>0$ for all $j$. We further split $V$ as $V=[V_1 ~~V_2]$ where $V_1\in\R^{N\times M}$ and $V_2\in\R^{N\times (N-M)}$. It is easy to show the following
\begin{equation}
\Psi\triangleq\Sigma^{-1}U^T \in \R^{M\times M}; 
~~V
\left [
\begin{array}{c}
\Psi A x\\
V^T_2 x\\
\end{array}
\right ]=x, \textrm{ for all } x\in \R^N.
\label{Psi}
\end{equation}
Define $\Phi:V^T_2\textrm{\LipSet}\to\R^M$ by
\begin{equation}
\Phi(V^T_2x)\triangleq Ax, \textrm{ for all }x\in\textrm{\LipSet}.
\label{Phi}
\end{equation}
We will show that $V^T_2\textrm{\LipSet}$ is $(\Phi^{-1},\omega)$-Lipschitz. Indeed, for $x^1,x^2\in$\LipSet
\begin{equation}
\begin{aligned}
\|V^T_2x^1-V^T_2x^2\| \leq &
\left \|
\left [
\begin{array}{c}
V^T_1 x^1\\
V^T_2 x^1\\
\end{array}
\right ]
-
\left [
\begin{array}{c}
V^T_1 x^2\\
V^T_2 x^2\\
\end{array}
\right ]
\right \|
=\|V^Tx^1-V^Tx^2\|
\overset{V\text{ is unitary}}{=}\|x^1-x^2\| \\
\overset{x^1,x^2\in\textrm{\LipSet}}{\leq}
& \omega \|Ax^1-Ax^2\| 
\overset{\textrm{\eqref{Phi}}}{=}
\omega \|\Phi(V^T_2x^1)-\Phi(V^T_2x^2)\|.
\\
\end{aligned}
\label{PhiLip}
\end{equation}

Since  $V^T_2 \Omega_{A,\omega,\epsilon} \subseteq V^T_2\textrm{\LipSet} \subseteq\R^{N-M}$ is finite and $(\Phi^{-1},\omega)$-Lipschitz according to \eqref{PhiLip}, Lemma~1 says that there is an $\omega\sqrt{N-M}$-Lipschitz hypothesis $\Ge:\R^M\to\R^{(N-M)}$ such that $\Ge(\Phi(V^T_2x))=V^T_2x$ for all $x\in\Omega_{A,\omega,\epsilon}$, and consequently from \eqref{Phi},
\begin{equation}
\Ge(Ax)=V^T_2x, \textrm{ for all }x\in\Omega_{A,\omega,\epsilon}.
\label{VT2}
\end{equation}
Note that the output space of the hypothesis $G_\epsilon$ in \eqref{VT2} has dimension $N-M$, instead of $N$ in Theorem~1.
Now define $R:\R^{M}\to\R^N$ by
\begin{equation}
R(y)\triangleq V
\left [
\begin{array}{c}
\Psi y \\
\Ge(y)\\
\end{array}
\right ], y\in\R^M.
\label{defineR}
\end{equation}
The following shows (i):
\begin{equation}
R(Ax)=V
\left [
\begin{array}{c}
\Psi  Ax \\
\Ge(Ax)\\
\end{array}
\right ] 
\overset{\textrm{\eqref{VT2}}}{=}V
\left [
\begin{array}{c}
\Psi Ax \\
V^T_2x\\
\end{array}
\right ]
\overset{\textrm{\eqref{Psi}}}{=}x,
\textrm{ for all } x\in \Omega_{A,\omega,\epsilon}.
\label{R}
\end{equation}
The following shows (iii):
\begin{equation}
A(R(Ax))=AV
\left [
\begin{array}{c}
\Psi  Ax \\
\Ge(Ax)\\
\end{array}
\right ] 
=U[\Sigma~~\textbf{0}]
\left [
\begin{array}{c}
\Sigma^{-1}U^T Ax \\
\Ge(Ax)\\
\end{array}
\right ]
=Ax,
\textrm{ for all } x\in \R^N.
\end{equation}
To show (ii), we note that similar to Proof of Theorem 1, for $x\in$\LipSet, there is an $x^j\in\Omega_{A,\omega,\epsilon}$, such that $Ax,Ax^j\in v_j$.
\begin{equation}
\begin{aligned}[r]
\|\Ge(Ax)-V^T_2x\|~~~~~\leq
&~\|\Ge(Ax)-V^T_2x^j\|+\|V^T_2x^j-V^T_2x\|\\
\overset{\textrm{\eqref{VT2}}}{=}
&~\|\Ge(Ax)-\Ge(Ax^j)\|+\|V^T_2x^j-V^T_2x\|\\
\overset{\textrm{\eqref{PhiLip}}}{\leq}
&~\|\Ge(Ax)-\Ge(Ax^j)\|+\omega\|Ax^j-Ax\|\\
\overset{\Ge\textrm{ is }\omega\sqrt{N-M}\textrm{-Lipschitz}}{\leq}
&~\omega\sqrt{N-M}\|Ax-Ax^j\|+\omega\|Ax^j-Ax\|\\
=&~\omega(1+\sqrt{N-M})\|Ax-Ax^j\|
\overset{Ax,Ax^j\in v_j,\textrm{\eqref{thm3}}}{\leq}
\epsilon, ~~x\in \textrm{\LipSet}.\\
\end{aligned}
\label{GVT2}
\end{equation}
Finally, for $x\in$\LipSet,
\begin{equation}
\begin{aligned}[c]
\|R(Ax)-x\|
\overset{\textrm{\eqref{defineR} and \eqref{Psi}}}{=}
&
\left \| V\left [
\begin{array}{c}
\Psi Ax\\\Ge(Ax)
\end{array}
\right ]
-
V\left [
\begin{array}{c}
\Psi Ax\\ V^T_2x
\end{array}
\right ]
\right \|
=
\left \|
\left [
\begin{array}{c}
0\\ \Ge(Ax)-V^T_2x
\end{array}
\right ]
\right \| \\
=
&\|\Ge(Ax)-V^T_2x\|
\overset{\textrm{\eqref{GVT2}}}
{\leq} 
\epsilon. ~~~~~~~~~~~~~~~~~~~~~~~~~~~Q.E.D.
\end{aligned}
\end{equation}


\newpage

\small
\bibliographystyle{plain}
\bibliography{bibfile}

\end{document}